\documentclass{article}
\usepackage{booktabs}
\usepackage{xcolor}
\usepackage{colortbl}
\usepackage{amssymb}
\usepackage{pifont}
\usepackage{graphicx}

\newcommand{\cmark}{{\color{green}\ding{51}}} % Green tick
\newcommand{\xmark}{{\color{red}\ding{55}}}   % Red cross

    \PassOptionsToPackage{numbers, compress}{natbib}
\usepackage[final]{neurips_2024}

\usepackage[utf8]{inputenc} % allow utf-8 input
\usepackage[T1]{fontenc}    % use 8-bit T1 fonts
\usepackage{hyperref}       % hyperlinks
\usepackage{cleveref}
\usepackage{url}            % simple URL typesetting
\usepackage{booktabs}       % professional-quality tables
\usepackage{amsfonts}       % blackboard math symbols
\usepackage{nicefrac}       % compact symbols for 1/2, etc.

\usepackage{microtype}
\usepackage{xcolor}         % colors
\title{Contamination Report for Multilingual Benchmarks}

\author{%
  Sanchit Ahuja$^*$ \\
  Microsoft Research \\
  \And
  Varun Gumma$^*$ \\
  Microsoft Research \\
  \And
  Sunayana Sitaram \\
  Microsoft Research \\
}

\begin{document}

\maketitle
\def\thefootnote{*}\footnotetext{Denotes equal contribution. Correspondence to \texttt{sunayana.sitaram@microsoft.com}}
\begin{abstract}
Benchmark contamination refers to the presence of test datasets in Large Language Model (LLM) pre-training or post-training data. Contamination can lead to inflated scores on benchmarks, compromising evaluation results and making it difficult to determine the capabilities of models. In this work, we study the contamination of popular multilingual benchmarks in LLMs that support multiple languages. We use the Black Box test to determine whether $7$ frequently used multilingual benchmarks are contaminated in $7$ popular open and closed LLMs and find that almost all models show signs of being contaminated with almost all the benchmarks we test. Our findings can help the community determine the best set of benchmarks to use for multilingual evaluation.
\end{abstract}
\section{Introduction}
Large Language Models (LLMs) have shown significant improvements on standard benchmarks as compared to their predecessors \cite{ahuja-etal-2023-mega,ahuja-etal-2024-megaverse}. These models are pre-trained on large amounts of data collected from the web via crawling, in which a significant portion of the Internet is consumed and often memorized by such large scale models during training. Such rampant data collection might unexpectedly capture publicly available benchmarks, causing LLMs to ingest test sets and memorize them, leading to a high score upon evaluation \cite{sainz-etal-2024-data}. This phenomenon is called \textit{data-contamination}, and it paints a false picture of the abilities of an LLM. LLMs also undergo an instruction-tuning phase, and are sometimes further tuned via RLHF, where the model is trained on task specific datasets. However, LLM creators do not always disclose the exact details of the datasets used, and it is plausible that the model is trained on benchmark datasets intentionally or unintentionally. Hence, \textit{contamination} can occur during the pre-training or post-training phases \cite{dubey2024llama3herdmodels}. In this work, we study the contamination of $7$ recent LLMs on $7$ popular multilingual benchmarks used in prior work to evaluate the capabilities of LLMs on non-English languages. Our main contribution in this paper is an analysis of which multilingual benchmark in contaminated in which model by utilizing the contamination detection technique proposed by \citet{oren2023proving}. 
\section{Related Works}

Various methods have been developed to identify dataset contamination for scenarios in which LLM training data is disclosed, as well as not disclosed. For example, \citet{yang2023rethinking}'s LLM Decontaminator quantifies rephrased samples by comparing them to a benchmark but needs access to training data. Other methods by \citet{oren2023proving} and \citet{golchin2023time} do not require training data; they use techniques like analyzing log probabilities of open source models or guided prompting. A recent survey by \citet{ravaut2024much} offers an extensive review of these strategies. \\
Some of the previous works in Multilingual Evaluation such as \citet{ahuja-etal-2023-mega} tries to tackle the problem of identifying contamination in GPT-4 by prompting the model to fill the dataset cards. Another work by \citet{ahuja-etal-2024-megaverse} follows the same method as ours albeit at a smaller scale.
\section{Methodology}
We follow the Black Box test for contamination detection in open-source models, as described by \citet{oren2023proving}. This test is a statistical approach that offers provable guarantees for identifying whether a given test set has been contaminated. The key idea behind this method is to exploit a property common to many datasets, known as \textit{exchangeability}. This property ensures that the joint distribution of the dataset remains unchanged regardless of the order in which the examples are presented.

If a model has been exposed to a benchmark dataset, it will likely develop a bias toward the canonical ordering of examples—the sequence in which they are presented in public repositories—over randomly shuffled versions of the same dataset. By comparing a model's performance on the canonical order versus shuffled orders, this method determines if the model exhibits a statistically significant preference for the original order. If such a difference is found, it provides evidence that the test set has been contaminated according to this framework.

In this reproduction study, we evaluate $7$ models (\textsc{Mistral-7B}, \textsc{Mistral-7B-it} \cite{jiang2023mistral7b}, \textsc{Llama-3.1-8B}, \textsc{Llama-3.1-8B-it} \cite{dubey2024llama3herdmodels}, \textsc{Gemma-2-9B}, \textsc{Gemma-2-9B-it} \cite{gemmateam2024gemma2improvingopen}, and \textsc{Aya-23-8B} \cite{aryabumi2024aya23openweight}) on $7$ multilingual datasets (\textsc{Xnli} \cite{conneau-etal-2018-xnli}, \textsc{XQuad} \cite{artetxe-etal-2020-cross}, \textsc{XStoryCloze} \cite{lin-etal-2022-shot}, \textsc{XCopa} \cite{ponti-etal-2020-xcopa}, \textsc{XLSum} \cite{hasan-etal-2021-xl}, \textsc{Flores} \cite{nllbteam2022languageleftbehindscaling}, \textsc{Paws-X} \cite{yang-etal-2019-paws}). The rationale behind evaluating both base and instruction tuned variants of a model is to understand, in which phase (pretraining or posttraining) contamination occurs. We use $5000$ data points overall uniformly spread across all the languages of the datasets, split across $48$ shards with $r=768$ permutations per shard. Hence, according to \citet{oren2023proving}, we have a significance value of $1/(1 + r) = 0.0013$, and any $p$-val lower than this threshold is considered as contamination. All these experiments were run on $8 \times$ H100s for $362$ hours ($\approx 15$ days).

\section{Results and Discussions}

Table \ref{tab:contamination} lists the datasets affected by contamination. We observe that only 4 instances show no contamination, while a significant portion of the datasets, which were not contaminated in previous versions of the models (Table \ref{tab:previous_results}), are now impacted. This indicates that newer versions LLMs, despite being larger and trained on more data, are more likely to include benchmark datasets in their training data. Given that the pre-training corpus for these models is typically expanded and reused, it is likely that future versions will also ingest these datasets. Our findings suggest that contamination occurs during the pre-training phase and persists after post-training.

\begin{table}[h]
\resizebox{\textwidth}{!}{%
\begin{tabular}{@{}l|ccccccc@{}}
\toprule
            & \textsc{Llama-3.1-8B} & \textsc{Llama-3.1-8B-it} & \textsc{Mistral-7B-v0.3} & \textsc{Mistral-7B-v0.3-it} & \textsc{Gemma-2-9b-it} & \textsc{Gemma-2-9b} & \textsc{Aya-23-8B} \\ \midrule
\textsc{Flores}      & \xmark  & \xmark  & \xmark  & \xmark  & \xmark  & \xmark  & \xmark \\
\textsc{Paws-X}      & \xmark  & \xmark  & \xmark  & \xmark  & \xmark  & \xmark  & \cmark \\
\textsc{XCopa}       & \xmark  & \xmark  & \xmark  & \xmark  & \xmark  & \xmark  & \cmark \\
\textsc{XLSum}       & \cmark  & \cmark  & \xmark  & \xmark  & \xmark  & \xmark  & \xmark  \\
\textsc{Xnli}        & \xmark  & \xmark  & \xmark  & \xmark  & \xmark  & \xmark  & \xmark  \\
\textsc{XQuad}       & \xmark  & \xmark  & \xmark  & \xmark  & \xmark  & \xmark  & \xmark  \\
\textsc{XStoryCloze} & \xmark  & \xmark  & \xmark  & \xmark  & \xmark  & \xmark  & \xmark  \\
\bottomrule
\end{tabular}%
 }
 \vspace{2mm}
\caption{Benchmark contamination presence in the evaluated models. \xmark\ means \textbf{contaminated} and \cmark\ means \textbf{not contaminated}.}
\label{tab:contamination}
\end{table}

\begin{table}[h]
\small 
\centering
\begin{tabular}{@{}l|ccc@{}}
\toprule
 & \textsc{Gemma-7B-it} & \textsc{Llama-2-7B-it} & \textsc{Mistral-7B-v0.1-it} \\ \midrule
\textsc{Paws-X}     & \xmark & \xmark & \xmark \\
\textsc{XCopa}      & \xmark & \xmark & \xmark \\
\textsc{Xnli}       & \cmark & \cmark & \cmark \\
\textsc{XQuad}     & \cmark & \xmark & \xmark \\
\textsc{XStoryCloze} & \cmark & \cmark & \cmark \\ \bottomrule
\end{tabular}
 \vspace{2mm}
\caption{Previous contamination results from \citet{ahuja-etal-2024-megaverse}. We use this table for cross-comparison.}
\label{tab:previous_results}
\end{table}

It is crucial to detect and prevent contamination, especially in multilingual datasets, which are both costly to create and relatively scarce. In future work, we aim to expand our analysis by evaluating a larger number of datasets and models for contamination. We hope our efforts will guide future research in carefully selecting benchmarks for multilingual evaluation.

\section{Limitations}
As the contamination evaluation is a computationally expensive process, we are constrained in the number of datasets and models we can evaluate. We choose the most popular benchmarks and models available publicly. Due to space constraints, we provide an initial analysis of the results, and future work can build upon this work. Further, the framework used only identifies if the dataset is contaminated for a certain model or not, and does not identify the extent of contamination, for which access to training data is required.

\bibliography{custom}

% \appendix
% \input{Styles/content/appendix}
\newpage
\section*{NeurIPS Paper Checklist}

\begin{enumerate}

\item {\bf Claims}
    \item[] Question: Do the main claims made in the abstract and introduction accurately reflect the paper's contributions and scope?
    \item[] Answer: \answerYes{}{} % Replace by \answerYes{}, \answerNo{}, or \answerNA{}.
    \item[] Justification: We propose this work as a lookup table to have an idea about which datasets are contaminated before using these datasets for downstream evaluations.

\item {\bf Limitations}
    \item[] Question: Does the paper discuss the limitations of the work performed by the authors?
    \item[] Answer: \answerYes{} % Replace by \answerYes{}, \answerNo{}, or \answerNA{}.
    \item[] Justification: We have added the limitations section in our paper above.

\item {\bf Theory Assumptions and Proofs}
    \item[] Question: For each theoretical result, does the paper provide the full set of assumptions and a complete (and correct) proof?
    \item[] Answer: \answerNA{}{} % Replace by \answerYes{}, \answerNo{}, or \answerNA{}.
    \item[] Justification: This is an empirical paper with no theoretical results or proof part of the paper.

    \item {\bf Experimental Result Reproducibility}
    \item[] Question: Does the paper fully disclose all the information needed to reproduce the main experimental results of the paper to the extent that it affects the main claims and/or conclusions of the paper (regardless of whether the code and data are provided or not)?
    \item[] Answer: \answerYes{}{} % Replace by \answerYes{}, \answerNo{}, or \answerNA{}.
    \item[] Justification: We use open-source codebase to run our experiments. We also disclose all the information needed to reproduce our results.

\item {\bf Open access to data and code}
    \item[] Question: Does the paper provide open access to the data and code, with sufficient instructions to faithfully reproduce the main experimental results, as described in supplemental material?
    \item[] Answer: \answerYes{} % Replace by \answerYes{}, \answerNo{}, or \answerNA{}.
    \item[] Justification: We will open-source our codebase post-acceptance of the paper.

\item {\bf Experimental Setting/Details}
    \item[] Question: Does the paper specify all the training and test details (e.g., data splits, hyperparameters, how they were chosen, type of optimizer, etc.) necessary to understand the results?
    \item[] Answer: \answerYes{} % Replace by \answerYes{}, \answerNo{}, or \answerNA{}.
    \item[] Justification: We have provided all the details about how we perform all the experiments.

\item {\bf Experiment Statistical Significance}
    \item[] Question: Does the paper report error bars suitably and correctly defined or other appropriate information about the statistical significance of the experiments?
    \item[] Answer: \answerNo{} % Replace by \answerYes{}, \answerNo{}, or \answerNA{}.
    \item[] Justification: We did not re-run the experiments multiple times to calculate standard deviation or error bars since the experiments were computationally expensive.

\item {\bf Experiments Compute Resources}
    \item[] Question: For each experiment, does the paper provide sufficient information on the computer resources (type of compute workers, memory, time of execution) needed to reproduce the experiments?
    \item[] Answer: \answerYes{} % Replace by \answerYes{}, \answerNo{}, or \answerNA{}.
    \item[] Justification: We mention the details in the paper.
    
\item {\bf Code Of Ethics}
    \item[] Question: Does the research conducted in the paper conform, in every respect, with the NeurIPS Code of Ethics \url{https://neurips.cc/public/EthicsGuidelines}?
    \item[] Answer: \answerYes{} % Replace by \answerYes{}, \answerNo{}, or \answerNA{}.
    \item[] Justification: Yes, the paper follows the NeurIPS code of ethics.

\item {\bf Broader Impacts}
    \item[] Question: Does the paper discuss both potential positive societal impacts and negative societal impacts of the work performed?
    \item[] Answer: \answerNA{} % Replace by \answerYes{}, \answerNo{}, or \answerNA{}.
    \item[] Justification: 
    
\item {\bf Safeguards}
    \item[] Question: Does the paper describe safeguards that have been put in place for responsible release of data or models that have a high risk for misuse (e.g., pretrained language models, image generators, or scraped datasets)?
    \item[] Answer: \answerNA{} % Replace by \answerYes{}, \answerNo{}, or \answerNA{}.
    \item[] Justification: Our work only involves studying contamination for open-source datasets.

\item {\bf Licenses for existing assets}
    \item[] Question: Are the creators or original owners of assets (e.g., code, data, models), used in the paper, properly credited and are the license and terms of use explicitly mentioned and properly respected?
    \item[] Answer: \answerYes{} % Replace by \answerYes{}, \answerNo{}, or \answerNA{}.
    \item[] Justification: We have cited their work in our paper.

\item {\bf New Assets}
    \item[] Question: Are new assets introduced in the paper well documented and is the documentation provided alongside the assets?
    \item[] Answer: \answerNA{} % Replace by \answerYes{}, \answerNo{}, or \answerNA{}.
    \item[] Justification: No new assets are being released.

\item {\bf Crowdsourcing and Research with Human Subjects}
    \item[] Question: For crowdsourcing experiments and research with human subjects, does the paper include the full text of instructions given to participants and screenshots, if applicable, as well as details about compensation (if any)? 
    \item[] Answer: \answerNA{} % Replace by \answerYes{}, \answerNo{}, or \answerNA{}.
    \item[] Justification: No humans were involved in this study.

\item {\bf Institutional Review Board (IRB) Approvals or Equivalent for Research with Human Subjects}
    \item[] Question: Does the paper describe potential risks incurred by study participants, whether such risks were disclosed to the subjects, and whether Institutional Review Board (IRB) approvals (or an equivalent approval/review based on the requirements of your country or institution) were obtained?
    \item[] Answer: \answerNA{} % Replace by \answerYes{}, \answerNo{}, or \answerNA{}.
    \item[] Justification: There is no research involved with human subjects.

\end{enumerate}

\end{document}